\documentclass[a4paper]{article}

\usepackage{arxiv}
\usepackage[ruled,vlined]{algorithm2e}
\usepackage{tabularx}
\usepackage{url} 
\usepackage{multirow}
\usepackage{makecell}
\usepackage{microtype}
\usepackage{color, colortbl}
\usepackage[table]{xcolor}
\usepackage[colorinlistoftodos, textwidth=20mm]{todonotes}
\usepackage{mathtools}
\usepackage{eucal}
\usepackage{booktabs}
\usepackage{amsfonts}

\newif\ifdraftydraft
\draftydrafttrue

\def\x{{\boldsymbol x}}
\def\param{{\boldsymbol \theta}}
\def\y{{\boldsymbol y}}
\def\yhat{\hat \y}

\def\tco{{\it train-clean-100}}
\def\tct{{\it train-clean-360}}
\def\tof{{\it train-other-500}}
\def\tsmall{{\it train-10h}}
\def\devclean{{\it dev-clean}}
\def\devother{{\it dev-other}}
\def\testclean{{\it test-clean}}
\def\testother{{\it test-other}}

\def\papername{slimIPL}
\definecolor{grey}{cmyk}{0.1,0.1,0.1,0.1}
\definecolor{orange}{cmyk}{0.1,0.2,0.4,0.0}

\newcommand{\librispeech}{{LibriSpeech}}
\newcommand{\librilight}{{Libri-Light}}

\title{\papername{}: Language-Model-Free Iterative Pseudo-Labeling}

\author{
    Tatiana Likhomanenko \quad 
    Qiantong Xu \quad 
    Jacob Kahn \quad 
    Gabriel Synnaeve \quad
    Ronan Collobert \\
    Facebook AI Research \\
    \texttt{antares@fb.com}
}
\begin{document}

\maketitle
\begin{abstract}
Recent results in end-to-end automatic speech recognition have demonstrated the efficacy of pseudo-labeling for semi-supervised models trained both with Connectionist Temporal Classification (CTC) and Sequence-to-Sequence (seq2seq) losses. 
Iterative Pseudo-Labeling (IPL), which continuously trains a single model using pseudo-labels iteratively re-generated as the model learns, has been shown to further improve performance in ASR. We improve upon the IPL algorithm: as the model learns, we propose to iteratively re-generate transcriptions with hard labels (the most probable tokens), that is, \textit{without} a language model. 
We call this approach Language-Model-Free IPL (\papername{}) and give a resultant training setup for low-resource settings with CTC-based models.
\papername{} features a dynamic cache for pseudo-labels which reduces sensitivity to changes in relabeling hyperparameters and results in improved training stability. \papername{} is also highly-efficient and requires 3.5-4x fewer computational resources to converge than other state-of-the-art semi/self-supervised approaches.
With only 10 hours of labeled audio, \papername{} is competitive with self-supervised approaches, and is state-of-the-art with 100 hours of labeled audio without the use of a language model both at test time and during pseudo-label generation. Moreover, slimIPL is applicable to conversational speech and its hyperparameters are transferred out-of-the-box.

\end{abstract}

\section{Introduction}
\label{sec:into}
Recent work in deep learning has shifted towards methods which can efficiently learn from large amounts of unlabeled data to improve performance and decrease costs associated with labeling. Semi-supervised learning~\cite{chapelle2006alexanderzien} combines information from both labeled and unlabeled data; the amount of unlabeled data typically exceeds the amount of labeled data. In automatic speech recognition (ASR), while many recent semi-supervised methods outperform a supervised baseline in a low-resource setting, a gap between semi- and fully-supervised training remains. Further, not all of the approaches are equally scalable as the amount of labeled and unlabeled data increases, as is the case in recent setups such as the \librilight{} benchmark~\cite{librilight}.

Some of the earliest and simplest semi-supervised approaches use self-training~\cite{scudder1965probability}. 
Self-training employs a base model trained with labeled data which acts as a ``teacher'' and is used to label unlabeled data (the resulting labels are referred as ``pseudo-labels'', PLs). A ``student'' model is then trained (typically from scratch) with both labeled and pseudo-labeled data to yield a final model. For competitive results in ASR, a language model (LM) was a key component of pseudo-labeling: it is usually combined with the acoustic model via beam-search decoding~\cite{hsu2020semi,xu2020iterative,xu2020self} or through shallow fusion~\cite{kahn2020self,synnaeve2019end,park2020improved,zhang2020pushing} to generate PLs. With this setting, however, acoustic models tend to over-fit to the text training set of the LM used for pseudo-labeling~\cite{xu2020iterative,park2020improved}.

In this work, we show that competitive pseudo-labeling approaches rely neither on beam-search decoding nor on a language model. In our setup, pseudo-labels are generated by picking hard labels -- tokens with the highest acoustic model probability.
Our approach is based on the recently-proposed iterative pseudo-labeling algorithm (IPL)~\cite{xu2020iterative}: we continuously train a single model using iteratively re-generated pseudo-labels as model learns. We call our algorithm language-model-free IPL (\papername{}) and give its overview in Section~\ref{sec:vpl}. We demonstrate in Section~\ref{sec:exp} that this approach is effective for models trained with Connectionist Temporal Classification (CTC)~\cite{graves2006connectionist} in low-resource settings and is competitive with current state-of-the-art results. Ablation studies on different aspects of the proposed algorithm in Section~\ref{sec:ablation} show \papername{} provides training stability and a robustness to hyperparameter settings.

\section{Related Work}
Self-training methods~\cite{scudder1965probability} still attract researchers: extensions to self-training are numerous and include (a) selecting particular subsets of pseudo-labeled data for student training, (b) the reiteration of the PL procedure several times to progressively-improve the teacher model, (c) the introduction of different types of noise for student model training, and (d) sampling techniques and schedules for training over labeled and pseudo-labeled datasets.
Many recent works on self-training propose and validate these extensions, including those in computer vision~\cite{yalniz2019billion,xie2020self}, natural language processing~\cite{yarowsky1995unsupervised,mcclosky2006effective,reichart2007self,huang2009self,he2020revisiting,ueffing2006using,zhang2016exploiting}, ASR~\cite{novotney2009analysis,kahn2020self,parthasarathi2019lessons,synnaeve2019end,park2020improved}, and speech translation~\cite{pino2020self}. 

One extension to the simple pseudo-labeling method consists of continuously training a single model~\cite{lee2013pseudo}. At the beginning of training, a model is trained only on labeled data after which training continues on data jointly-selected from both labeled and unlabeled datasets. PL re-generation occurs after some number of iterations, and a supervised loss is computed both on labeled and pseudo-labeled data for each batch. An additional parameter determines the contribution of pseudo-labeled data to the overall loss. The effectiveness of this iterative training for a single model has been validated on tasks in vision~\cite{arazo2019pseudo}, natural language processing~\cite{he2020revisiting}, and ASR~\cite{chen2020semi,xu2020iterative}. 
Below, we give an overview of the most relevant approaches in ASR to our work.
\\
{\bf Iterative pseudo-labeling (IPL)~\cite{xu2020iterative}}~algorithm follows prior work~\cite{lee2013pseudo} and uses augmentation of both labeled and unlabeled data, and continuously trains a single model with iterative re-generation of PLs by beam-search decoding with an LM, as the model learns. Compared to IPL, we maintain a dynamic cache with PLs and don't use any beam-search decoding or an LM.
\\
{\bf Noisy self-training~\cite{park2020improved}} performs five iterations of student network training, each time from scratch, with PLs generated by a teacher network.
In this approach, as is the case with IPL, shallow fusion with an LM is used with a decoding procedure to generate PLs, while \papername{} doesn't use an LM at all.
\\
{\bf Self-training~\cite{chen2020semi}}~is the closest to our work: the authors continuously train a model with re-generated hard PLs after {\it each} iteration. This work is more focused on studying the impact of noise, and both SpecAugment~\cite{park2019specaug} and speed perturbation are applied for labeled and unlabeled data during training. Our experiments show that re-generating PLs with hard labels after {\it each} iteration causes training instability resulting in divergence, whereas \papername{} exploits a dynamic cache mechanism to stabilize training.
\\
{\bf wav2vec~\cite{baevski2020wav2vec}}'s~unsupervised pre-training gives a significant boost in performance for low-resource settings. Training has two steps: first, pre-training on unlabeled data by masking the input audio in the latent space and solving a contrastive learning task~\cite{oord2018representation}; second, fine-tuning the model using labeled audio only. Recently, one-step training~\cite{talnikar2020joint} is proposed to directly optimize a downstream task which simplifies hyperparameter tuning. One of the known problems with contrastive training is a need of large batches~\cite{baevski2020wav2vec,talnikar2020joint}. 

\section{Pseudo-Labeling}
\label{sec:pl}
Let $L = \{\x_i, \y_i\}$ be a labeled dataset and $U = \{\x_j\}$ a large unlabeled dataset. 
We consider a semi-supervised pseudo-labeling approach where the acoustic model (AM) $\mathcal{M}_\theta$ is continuously trained on combination of a labeled set and an iteratively re-generated pseudo-labeled set.
Training minimizes the following loss function: $\mathcal{L}(\param) = \mathcal{L}_L(\param) + \lambda \mathcal{L}_U(\param), \, \lambda\in\mathbb{R}^+$, where $\param$ are the parameters of the AM, and $\lambda$ is a tunable parameter controlling the importance of unlabeled data. The losses for labeled data $\mathcal{L}_L$ and for unlabeled data $\mathcal{L}_U$ are defined as $\mathcal{L}_L(\param) = - \displaystyle \mathbb{E}_{\x,\y \sim p(\x,\y)} \log p_{\param}(\y|\x), \, (\x, \y)\in L,$ where $p(\x, \y)$ is the empirical data distribution of samples from $L$, $p_\param(\y|\x)$ is the conditional distribution defined by $\mathcal{M}_\theta$, $\mathcal{L}_U(\param) = - \displaystyle \mathbb{E}_{\x \sim p(\x)}  \log  p_{\param}(\hat{\y} |\x), \, \x\in U,$ where $p(\x)$ is the empirical data distribution of samples from $U$, and $\hat{\y}$ are the pseudo-labels for utterance~$\x \in U$.

One key difference in existing pseudo-labeling approaches is how the labels assignments $\hat{\y}$ are obtained for unlabeled data $\x\in U$. In the general literature, pseudo-labeling refers to the hard label generation:
\begin{equation}
\label{eq:greedy}
\hat \y = \underset{\y}{\operatorname{argmax}} ~ \log p_{\param}(\y|\x).
\end{equation}
In machine translation and ASR domains, the model $p_{\param}(\y|\x)$ is often sequence-to-sequence, and the solution of Eq.~\eqref{eq:greedy} may be approximated with a beam-search decoding algorithm~\cite{sennrich2016improving,he2020revisiting,synnaeve2019end,kahn2020self,xu2020iterative,park2020improved,pino2020self,chen2020semi}. Indeed, most recent work in ASR relies on an LM $p^{lm}(\y)$ to generate PLs, and instead optimizes:
\begin{equation}
\label{eq:greedylm}
\hat \y = \underset{\y}{\operatorname{argmax}} ~ \log p_{\param}(\y|\x) + \alpha \, \log p^{lm}(\y),\,\x\in U,
\end{equation}
where $\alpha$ is a hyperparameter controlling the amount of language model regularization.
Pseudo-labeling is also popular in computer vision~\cite{lee2013pseudo,sohn2020fixmatch}. Variants exist, such as ``soft labels'' $\hat \y = p_{\param}(\y|\x)$, variations on soft labeling~\cite{berthelot2019mixmatch,berthelot2019remixmatch}, "hard distillation"~\cite{touvron2020training} and sampling~\cite{imamura2018enhancement,edunov2018understanding}.

\setlength{\textfloatsep}{5pt}
\begin{algorithm}[t!]
\caption{\papername{}}
\label{algo:ipl}
\SetAlgoLined
\KwData{labeled $L = \{\x_i, \y_i\}$ and unlabeled $U = \{\x_j\}$}
\KwResult{Acoustic model $\mathcal{M}_\theta$}
    1. Train $\mathcal{M}_\theta$ on $L$ with augmentation for $M$ updates\;
    2. \While{cache is not full at size $C$}{
        - Draw a random batch from $\x \in U$\;
        - Generate its PL $\hat \y$ by $\mathcal{M}_\theta$ following Eq.(\ref{eq:greedy})\;
        - Store $\{\x, \hat \y\}$ into the cache\;
        - Train $\mathcal{M}_\theta$ on $L$ with augmentation for $1$ update\;
        }
    3. Decrease model's $\mathcal{M}_\theta$ dropout\;

  \Repeat{convergence or maximum iterations are reached}{
    4. Train $\mathcal{M}_\theta$ on $L$ with augmentation for $N_L$ updates\;
    5. \For{$N_{U}$ updates}{
        - Draw a random batch $B = \{\x, \hat \y\}$ from the cache\;
        - {\bf With probability $p$}, $B$ is removed from cache and a new pair of random batch $\x' \in U$\ and its PL $\hat \y'$ generated by $\mathcal{M}_\theta$ is added in\;
        - Apply augmentation to batch $B$ and make an optimization step to update $\mathcal{M}_\theta$.
    }
  }
\end{algorithm}
\section{Language-Model-Free IPL}
\label{sec:vpl}

In the original IPL algorithm~\cite{xu2020iterative}, PLs are generated with a beam-search decoder leveraging an LM and approximating the solution of Eq.~\eqref{eq:greedylm}. While the main motivation is to transfer the knowledge of the LM into the AM, drawbacks exist: (i)~generating PLs is computationally intensive, and (ii)~models easily over-fit to LM knowledge. Regularization techniques are proposed in~\cite{xu2020iterative} to overcome (ii), such that one can still benefit from the LM when decoding at evaluation time.

We demonstrate that PLs do not need to rely on any LM information at all. Algorithm~\ref{algo:ipl} describes \papername{}. While it follows the IPL algorithm, PLs are generated by considering the top prediction according to the AM as per Eq.~(\ref{eq:greedy}). For CTC-based acoustic models, this corresponds exactly to choosing the most likely token at each time step. Our approach also imitates self-training~\cite{chen2020semi}, but instead of re-generating PLs after {\it each} update we exploit a {\it dynamic cache} to use PLs generated by the {\it previous model states} (this can be viewed as models ensemble averaging for PL generation). This stabilizes the optimization process and avoids sudden model divergence as discussed in Section~\ref{sec:ablation}. In addition, a regularization scheme is implemented via data augmentation over the input (acoustic) data, both for labeled~$L$ and unlabeled~$U$ samples. slimIPL has several hyperparameters: (i) when PL generation begins $M$, (ii) the proportion of labeled and unlabeled data $\lambda = N_U / N_L$, (iii) the dynamic cache size $C$ and the probability $p$ of updating the cache.

\begin{table*}[t!]
\caption{
WER comparison of our supervised baselines with prior work: LL-10 (top) and LS-100 (bottom).
\label{tab:resuts-sup}}
\begin{footnotesize}
\begin{center}
\setlength\tabcolsep{5pt} 
\begin{tabular}{@{}ccccccccc@{}}
\toprule
\multirow{2}{*}{Method} & \multirow{2}{*}{Stride} & \multirow{2}{*}{Tokens} & \multirow{2}{*}{Criterion} & \multirow{2}{*}{LM} & \multicolumn{2}{c}{Dev WER} & \multicolumn{2}{c}{Test WER} \\
\cmidrule(lr){6-7} \cmidrule(lr){8-9}
 &  & & & & clean        & other        & clean    & other        \\
\midrule
 \multirow{1}{*}{Libri-Light~\cite{librilight}} &  \multirow{1}{*}{20 ms} & \multirow{1}{*}{letters} & CTC & word 4-gram & 34 & 60.9 & 33.5 & 62.1 \\
\midrule
\multirow{3}{*}{Ours} & \multirow{3}{*}{30ms} & \multirow{3}{*}{letters} & \multirow{3}{*}{CTC} & - & 31.9 & 52.3 & 32.6 & 52.4 \\
 & & & & word 4-gram & 18.8 & 39.3 & 19.6 & 39.7 \\
 & & & & \, \, + rescoring & 17.1 & 38.2 & 17.9 & 38.9  \\
\midrule
\midrule
 \multirow{1}{*}{RWTH~\cite{luscher2019rwth}} & \multirow{1}{*}{-} & \multirow{1}{*}{-} & hybrid & word 4-gram & 5.0 & 19.5 & 5.8 & 18.6 \\
\midrule
 DeCoAR~\cite{ling2020deep} & - & phn. & CTC & - & - & - & 6.1 & 17.4 \\
\midrule
 Improved T/S~\cite{park2020improved} & - & 16k wp & S2S & - & 5.3 & 16.5 & 5.5 & 16.9 \\
\midrule
 \multirow{3}{*}{Ours} & \multirow{3}{*}{30ms} & \multirow{3}{*}{letters} & \multirow{3}{*}{CTC} & - & 6.2 & 16.8 & 6.2 & 16.8 \\
 & & & & word 4-gram & 4.1 & 12.4 & 4.5 & 12.7 \\
 & & & & \, \, + rescoring & 3.3 & 10.9 & 3.7 & 11.4  \\
 \bottomrule
\end{tabular}
\end{center}
\end{footnotesize}
\end{table*}

\section{Experiments}
\label{sec:exp}
\subsection{Data}
\label{sec:data}
All experiments are performed on the \librispeech{} dataset~\cite{panayotov2015librispeech} (contains 960 hours of training audio with paired transcriptions: \tco{}, \tct{}, and \tof{} parts) and \librilight{}~\cite{librilight} labeled limited resource training subset \tsmall{} originally extracted from \librispeech{}. We consider two low-resource scenarios with different amounts of labeled / unlabeled data: (i) LL-10/LS-960 uses \tsmall{} as labeled data and full \librispeech{} as unlabeled; (ii) LS-100/LS-860 as labeled data and \tct{} and \tof{} as unlabeled.
The standard \librispeech{} validation sets (\devclean{} and \devother{}) are used to tune all hyperparameters, as well as to select the best models. Test sets (\testclean{} and \testother{}) are used only to report final word error rate (WER) performance.
We keep the original 16kHz sampling rate and compute log-mel filterbanks with 80 coefficients for a 25ms sliding window, strided by 10ms. All features are normalized to have zero mean and unit variance per input sequence before feeding them into the acoustic model.

\subsection{Acoustic Models}
\label{sec:acoustic}
We consider CTC-based models. Architectures follow~\cite{synnaeve2019end}: the encoder is composed of a 1-D convolution with kernel size 7 and stride 3 followed by 36 4-head Transformer blocks~\cite{vaswani2017attention}. The self-attention dimension is $768$ and the feed-forward network (FFN) dimension is $3072$ (with 4 heads) in each Transformer block. The output of the encoder is followed by a linear layer to the output classes. We use dropout after the convolution, dropout on the self-attention and on the FFN for all Transformer layers, and layer drop~\cite{fan2019reducing}, dropping entire layers at the FFN level.
\\
{\bf Tokens}~~~Letters are used for all experiments. The letter set consists of the 26 English alphabet letters augmented with the apostrophe and a word boundary token.
\\
{\bf Data augmentation}~~~Training is performed with SpecAugment~\cite{park2019specaug} only. We use two frequency masks with frequency mask parameter $F=30$, ten time masks with maximum time-mask ratio $p=0.1$ and time mask parameter $T=50$; time warping is not used. For LL-10/LS-960 we found that twenty time masks with $T=25$ improve performance. \\
{\bf Training}~~~For all experiments we use the Adagrad optimizer~\cite{duchi2011adaptive} and decay learning rate by 2 each time the WER reaches a plateau on the validation sets. All models architectures, as well as \papername{} are implemented within the flashlight\footnote{ \url{https://github.com/flashlight/flashlight}} framework and available at~\url{https://github.com/flashlight/wav2letter/tree/master/recipes/slimIPL}. Models are trained with dynamic batching (effective average batch size is 14 per GPU) and mixed-precision computations on 16 GPUs (Volta 32GB) for 350-500k updates.

\subsection{Beam-search Decoding and Rescoring}
\label{sec:decoding}

In all our experimental results, we report not only WER without an LM, but also WER obtained with a one-pass beam-search decoder leveraging an LM. Following the notation introduced in Section~\ref{sec:pl}, the beam-search decoder aims at maximizing:
\begin{equation*}
\log p_{\param}(\yhat | \x) + \alpha \log p^{lm}(\yhat) + \beta|\yhat|,
\end{equation*}
where $\alpha$ and $\beta$ are hyperparameters to tune.
We rely on the beam-search decoder from the flashlight framework following~\cite{collobert2016wav2letter}: the lexicon-based beam-search decoder with a word-level LM. \librispeech{} validation sets, \devclean{} and \devother{}, are used to optimize the beam-search decoder hyperparameters, through random search. We also report WER obtained by rescoring the beam of hypothesis generated by the one-pass decoder. Rescoring is performed with a strong word-level Transformer LM, following the procedure described in~\cite{synnaeve2019end}. We use open-sourced word-level LMs trained on the \librispeech{} LM corpus: 4-gram~\cite{likhomanenko2019needs} and Transformer~\cite{synnaeve2019end} LMs.

\subsection{Supervised Baselines}
\label{sec:supervised}

The dropout and layer drop parameters of our acoustic models were set to 0.5 (0.3) when trained on LL-10 (LS-100).
Performance in WER is reported in Table~\ref{tab:resuts-sup}. Our supervised baseline models trained on either LL-10 or LS-100 define a new state-of-the-art both on \testclean{} and \testother{} with beam-search decoding and further rescoring. On \testother{} these models are state-of-the-art even without an LM.

\subsection{Semi-Supervised Experiments}
\begin{table*}[t!]
\caption{
Comparison with other semi- and unsupervised methods: LL-10/LS-960 (top) and LS-100/LS-860 (bottom).
\label{tab:resuts}}
\begin{center}
\setlength\tabcolsep{5pt} 
\resizebox{\linewidth}{!}{
\begin{tabular}{@{}cccccccccccc@{}}
\toprule
\multirow{2}{*}{Method} & \multirow{2}{*}{Stride} & \multirow{2}{*}{Tokens} & \multirow{2}{*}{Criterion} & \multirow{2}{*}{LM} & \multicolumn{2}{c}{Dev WER} & \multicolumn{2}{c}{Test WER} & \multicolumn{3}{c}{Compute Resources} \\
\cmidrule(lr){6-7} \cmidrule(lr){8-9} \cmidrule(lr){10-12}
 &  & & & & clean        & other        & clean    & other  & Train Time (Days) & \# G/TPUs & G/TPU-days \\
\midrule
 \multirow{1}{*}{Libri-Light~\cite{librilight}} & \multirow{1}{*}{20 ms} & \multirow{1}{*}{letters} & CTC & word 4-gram & 30.5 & 55.8 & 30.1 & 57.2 & - & - & - \\
\midrule
\multirow{2}{*}{IPL~\cite{xu2020iterative}} & \multirow{2}{*}{80ms} & \multirow{2}{*}{5k wp} & \multirow{2}{*}{CTC} & - & 23.8 & 25.7 & 24.6 & 26.5 & \multirow{2}{*}{3} & \multirow{2}{*}{64~GPUs} & \multirow{2}{*}{192}\\
 & & & & \, \, + rescoring & 23.5 & 25.5 & 24.4 & 26.0 & & \\
\midrule
\multirow{3}{*}{wav2vec 2.0~\cite{baevski2020wav2vec}} & \multirow{3}{*}{20ms} & \multirow{3}{*}{letters} & \multirow{3}{*}{CTC} & - & 8.1 & 12.0 & 8.0 & 12.1 & \multirow{3}{*}{2.3} & \multirow{3}{*}{128~GPUs} & \multirow{3}{*}{294.4} \\
 & & & & word 4-gram & 3.4 & 6.9 & 3.8 & 7.3 & & \\
 & & & & word Transf. & 2.9 & 5.7 & 3.2 & 6.1  & & \\
\midrule
  \multirow{3}{*}{slimIPL} & \multirow{3}{*}{30ms} & \multirow{3}{*}{letters} & \multirow{3}{*}{CTC} & - & 11.4 & 14 & 11.4 & 14.7 & \multirow{3}{*}{4.7} & \multirow{3}{*}{16~GPUs} & \multirow{3}{*}{75.2} \\
 & & & & word 4-gram & 6.6 & 9.6 & 6.8 & 10.5 && \\
 & & & & \, \,  + rescoring & 5.3 & 7.9 & 5.5 & 9.0  & & \\
\midrule
\midrule
 \multirow{2}{*}{IPL~\cite{xu2020iterative}} &  \multirow{2}{*}{80ms} & \multirow{2}{*}{5k wp} & \multirow{2}{*}{CTC} & - & 5.5 & 9.3 & 6.0 & 10.3 & \multirow{2}{*}{3} & \multirow{2}{*}{64~GPUs} & \multirow{2}{*}{192} \\
 & & & & \, \, + rescoring & 5.0 & 8.0 & 5.6 & 9.0 \\
\midrule
 \multirow{2}{*}{Improved T/S~\cite{park2020improved}} &  \multirow{2}{*}{-} & \multirow{2}{*}{16k wp} & \multirow{2}{*}{S2S} & - & 4.3 & 9.7 & 4.5 & 9.5 & \multirow{2}{*}{10 $\times$ 5} & \multirow{2}{*}{32~TPUs} & \multirow{2}{*}{1600} \\
  & & & & LSTM & 3.9 & 8.8 & 4.2 & 8.6 \\
\midrule
 \multirow{3}{*}{wav2vec~2.0~\cite{baevski2020wav2vec}} &  \multirow{3}{*}{20ms} & \multirow{3}{*}{letters} & \multirow{3}{*}{CTC} & - & 4.6 & 9.3 & 4.7 & 9.0  & \multirow{3}{*}{2.3} & \multirow{3}{*}{128~GPUs} & \multirow{3}{*}{294.4} \\
 & & & & word 4-gram & 2.3 & 5.7 & 2.8 & 6.0  \\
 & & & & word Transf. & 2.1 & 4.8 & 2.3 & 5.0 \\
\midrule
 \multirow{3}{*}{slimIPL} & \multirow{3}{*}{30ms} & \multirow{3}{*}{letters} & \multirow{3}{*}{CTC} & - & 3.7 & 7.3 & 3.8 & 7.5  & \multirow{3}{*}{5.2} & \multirow{3}{*}{16~GPUs} & \multirow{3}{*}{83.2} \\
 & & & & word 4-gram & 2.8 & 5.6 & 3.1 & 6.1 \\
 & & & & \, \, + rescoring & 2.2 & 4.6 & 2.7 & 5.2  \\
 \bottomrule
\end{tabular}
}
\end{center}
\end{table*}

\papername{} architectures are identical to their supervised counterparts, except for their dropout and layer drop values which are decreased after $M$ (supervised-only) updates to 0.1 for both settings, see Algorithm \ref{algo:ipl} step 3. 
Stronger regularization (via high dropout) is critical to avoid over-fitting when training with labeled-only data. These regularization parameters are then decreased to ``increase'' model capacity, as more data is involved during the semi-supervised training (see ablation in Table~\ref{tab:ablations} when dropout is not changed during the training).
slimIPL hyperparameters were tuned by performing a search over the following ranges: $M$ in 5k, 10k, 20k; $N_U$:$N_L$ in 1:1, 1:2, 2:1, \{3,4,5\}:1, 10:1, 20:1; $C$ in 10, 100, 1k; $p$ in 0.1, 0.5, 1. The best configuration results are presented in Table~\ref{tab:resuts}. On the LL-10/LS-960 setup, slimIPL is the best performing pseudo-label-based technique, on both \testclean{} and \testother{}, almost closing the gap with wav2vec~\cite{baevski2020wav2vec}. On LS-100/LS-860, we define a new state-of-the-art for LM-free approaches on both \testclean{} and \testother{}, while being similar to wav2vec with additional LM decoding.

\subsection{Ablations}
\label{sec:ablation}

\begin{figure}[!h]
\centering
    \includegraphics[width=8cm]{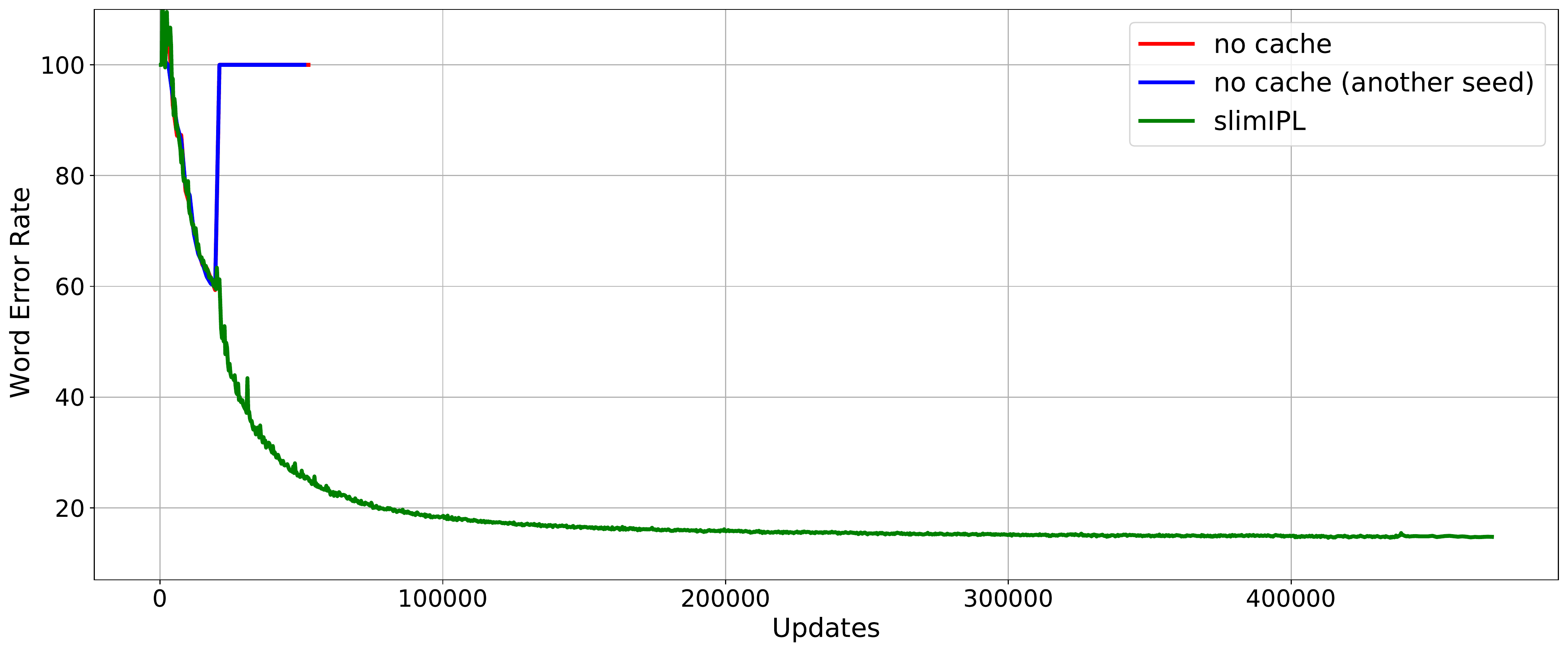}
    \includegraphics[width=8cm]{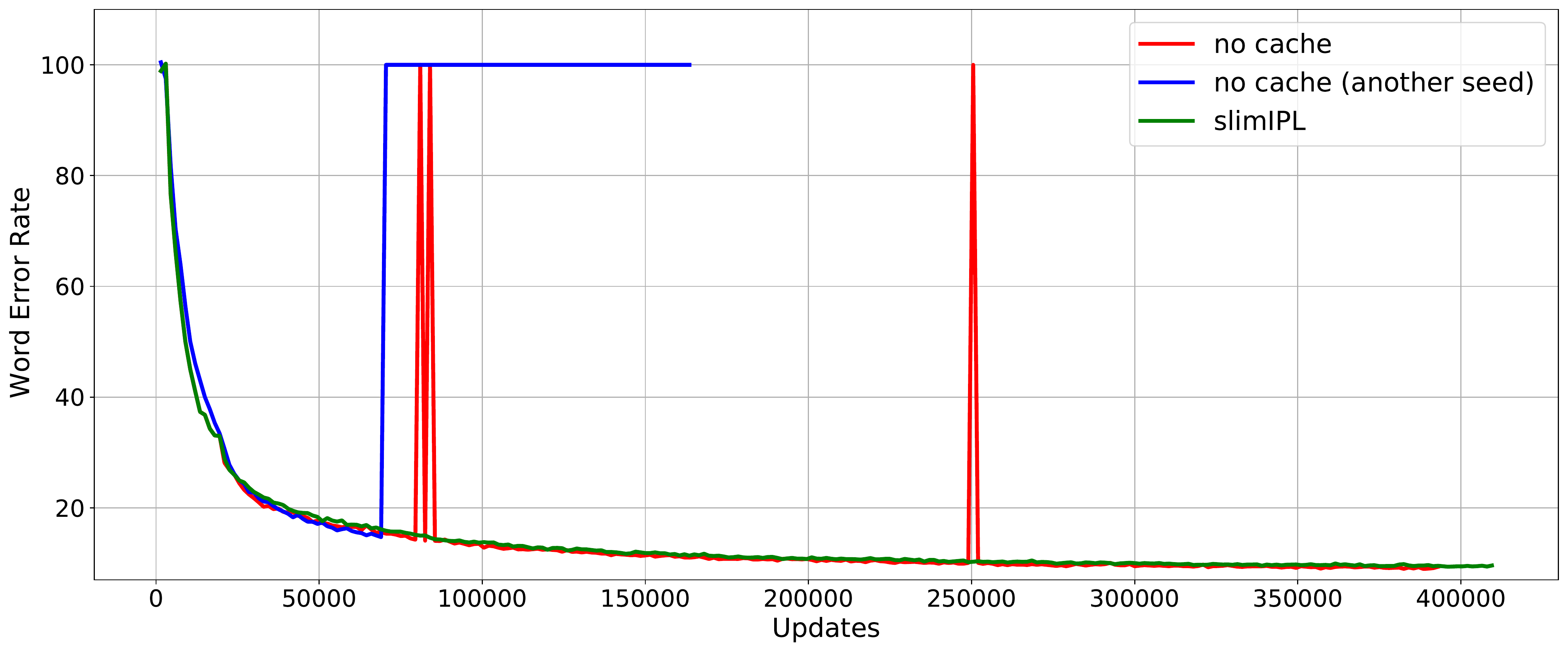}
\caption{Learning curves on {\it dev-other} for models trained on LL-10/LS-960 (left) and LS-100/LS-860 (right). slimIPL models refer to baseline models (grey) from Table~\ref{tab:ablations}.}
\label{fig:divergence}
\end{figure}

\begin{table}[t!]
\caption{
 Ablations study on slimIPL hyperparameters reporting validation WER: LL-10/LS-960 (top) and LS100/LS-860 (bottom). "x" refers to the values from a baseline (grey) model.
\label{tab:ablations}}
\begin{footnotesize}
\begin{center}
\setlength\tabcolsep{5pt} 
\begin{tabular}{@{}ccccccccc@{}}
\toprule
\multirow{2}{*}{dropout} & \multirow{2}{*}{$C$} & \multirow{2}{*}{$p$} & \multirow{1}{*}{$M/$} & \multirow{2}{*}{$\lambda$} & \multicolumn{2}{c}{no LM} & \multicolumn{2}{c}{4-gram LM} \\
\cmidrule(lr){6-7} \cmidrule(lr){8-9}
& & & WER-other & & clean & other & clean & other \\
\midrule
  \rowcolor{grey}
 0.5$\rightarrow$0.1 & 1000 & 0.1 & 20k/60 & 10/1 & 11.4 & 14.0 & 6.6 & 9.6  \\
 \midrule
 0.5 & x & x & x & x & 12.0 & 17.7 & 6.9 & 11.7  \\
 \midrule
 x & 10 & x & x & x & 17.8 & 20.7 & 9.0 & 12.8   \\
 x & 100 & x & x & x & 13.0 & 15.3 & 7.4 & 10.0 \\
 \midrule
 x & x & 0.5 & x & x & 12.3 & 15.8 & 7.0 & 10.1  \\
 x & x & 1 & x & x & 13.8 & 17.5 & 7.2 & 10.7 \\
 \midrule
 x & x & x & 5k/91 & x & 54.3 & 56.8 & 30.1 & 35.0 \\
 x & x & x & 10k/73 & x & 23.7 & 26.5 & 9.6 & 13.8 \\
 x & x & x & 40k/55 & x & 10.7 & 13.7 & 6.7 & 9.8  \\
 \midrule
 x & x & x & x & 1/1 & 11.3 & 15.0 & 6.6 & 10.3 \\ 
 x & x & x & x & 5/1 & 11.6 & 14.6 & 6.8 & 9.7  \\
 x & x & x & x & 20/1 & 11.9 & 14.7 & 6.5 & 9.8  \\
 \midrule
 \midrule
 \rowcolor{grey}
 0.3$\rightarrow$0.1 & 100 & 0.1 & 20k/33 & 1/1 & 3.6 & 7.5 & 2.9 & 5.9  \\
 \midrule
 0.3 & x & x & x & x & 4.1 & 8.1 & 3.1 & 6.3  \\
 \midrule
 x & 10 & x & x & x & 3.9 & 7.7 & 2.9 & 5.9  \\
 x & 1000 & x & x & x & 3.8 & 7.5 & 2.8 & 5.9  \\
 \midrule
 x & x & 0.5 & x & x & 3.8 & 7.3 & 3.0 & 5.9  \\
 x & x & 1 & x & x & 4.1 & 8.1 & 3.1 & 6.1 \\
 \midrule
 x & x & x & 5k/80 & x & 3.9 & 7.8 & 2.9 & 5.9  \\
 x & x & x & 10k/53 & x & 3.9 & 7.6 & 2.9 & 5.9  \\
  x & x & x & 50k/23 & x & 3.9 & 7.8 & 2.9 & 5.9  \\
 \midrule
 x & x & x & x & 1/2 & 4.2 & 8.5 & 3.1 & 6.4  \\
 x & x & x & x & 2/1 & 3.7 & 7.3 & 2.7 & 5.6  \\
 x & x & x & x & 3/1 & 3.7 & 7.3 & 2.8 & 5.6  \\
  x & x & x & x & 4/1 & 3.7 & 7.3 & 2.7 & 5.5  \\
\bottomrule
\end{tabular}
\end{center}
\end{footnotesize}
\end{table}

\begin{table}[t!]
\caption{
 Ablations study on slimIPL reporting validation WER: LL-10/LS-960 (top) and LS100/LS-860 (bottom). "Naive" approach~\cite{chen2020semi} is referred as $C=$"no"; slimIPL baseline (grey) models are the same as baseline models from Table~\ref{tab:ablations}.
\label{tab:ema}}
\begin{footnotesize}
\begin{center}
\setlength\tabcolsep{5pt} 
\begin{tabular}{@{}cccccc@{}}
\toprule
\multirow{2}{*}{EMA} & \multirow{2}{*}{Cache} & \multicolumn{2}{c}{no LM} & \multicolumn{2}{c}{4-gram LM} \\
\cmidrule(lr){3-4} \cmidrule(lr){5-6}
 & & clean & other & clean & other \\
\midrule
\rowcolor{grey}
 no & yes & 11.4 & 14.0 & 6.6 & 9.6  \\
 no & no & \multicolumn{4}{c}{diverges} \\
 \midrule
 yes & yes & 11.6 & 14.7 & 6.4 & 9.6  \\
 yes & no & 13.0 & 15.0 & 6.7 & 9.7 \\
 \midrule
 \midrule
 \rowcolor{grey}
 no & yes & 3.6 & 7.5 & 2.9 & 5.9  \\
 no & no & 3.9 & 7.5 & 3.1 & 5.9 \\
 \midrule
 yes & yes & 3.9 & 7.9 & 2.7 & 5.8  \\
 yes & no & 3.9 & 7.9 & 2.9 & 5.9 \\
\bottomrule
\end{tabular}
\end{center}
\end{footnotesize}
\end{table}

In Tables~\ref{tab:ablations} and ~\ref{tab:ema},
we study the robustness of slimIPL with respect to the choice of its hyperparameters. First, we consider the pseudo-labeling algorithm from~\cite{chen2020semi}, where PLs are generated from the acoustic model without dropout, from the acoustic samples without data augmentation, after {\it each} update (no cache, $C=no$). We found in practice that this approach is unstable, prone to unpredictable model divergence (generated transcriptions become empty), which can be occasionally recovered, see Figure~\ref{fig:divergence}.
In contrast, slimIPL is robust thanks to its dynamic cache strategy -- in practice we observed no divergence if the cache size $C\geq10$. For runs where the "cache-free" approach converges, we observe similar 
performance with slimIPL, as shown in Table~\ref{tab:ema}.
In addition, lowering the cache update probability $p$ allows to re-generate PLs more rarely in slimIPL than in cache-free approaches, which leads to faster training. slimIPL is robust to the cache size $C$ and cache update probability $p$. However, for limited supervision the small cache setting $C=10$ performs worse.

The starting update $M$ for involving unlabeled data in the training process is critical for the low-resource labeled data setting LL-10/LS-960. In the case of LS-100/LS-860, slimIPL recovers even when starting from high WER supervised models. To prevent quick over-fitting over supervised data for LL-10/LS-960, the ratio $\lambda$ between the number of unsupervised and supervised updates should be greater than $1$, with little variations in WER for any $\lambda > 1$. For LS-100/LS-860 this hyperparameter is less critical, $\lambda=4/1$ found to be optimal, and slower convergence observed for $\lambda < 1$.

Overall, with enough labeled data, slimIPL is robust to hyperparameter changes. When labeled data is limited, $C$, $M$ and $\lambda$ should be large enough to avoid over-fitting to labeled data.

\subsection{Exponential Moving Average}
slimIPL performs a kind of model averaging, as it leverages PL predictions from past versions of the acoustic model during training. This not only stabilizes the training (compared to only using predictions from the current model), but also is computationally efficient, thanks to the caching strategy. Another popular way of performing model averaging is by performing an exponential moving average (EMA) over past model weights, as training goes. We thus performed a comparison of slimIPL with a modified version of the algorithm in~\cite{chen2020semi}, where PLs are generated with the EMA acoustic model (without dropout, and without data augmentation over acoustic utterances). EMA decay factor is set to 0.999 to emulate the history range from the dynamic cache in our slimIPL experiments. In Table~\ref{tab:ema}, we show that the EMA approach stabilizes the training: no divergence is observed for the LL-10/LS-960 setup, and the model converges even if we regenerate PLs after {\it each} update. Combining the dynamic cache strategy and EMA does not seem to lead to improvements for both LL-10/LS-960 and LL-100/LS-860. Overall, slimIPL caching approach is as effective as an EMA approach, but with the advantage of being more memory and computationally efficient, as it saves the estimation of cached PLs and doesn't store additional EMA model. 

We perform EMA experiments only as ablation study while the concurrent work~\cite{manohar2021kaizen} performs in-depth study and analysis of EMA approach. In~\cite{manohar2021kaizen} authors propose more general framework with EMA, called Kaizen, which can be seen as a continuous version of the IPL for semi-supervised training. Authors also demonstrate that Kaizen with some range of EMA hyper-parameters is able to stabilize the training and improve overall performance.

\subsection{Efficiency}
\label{sec:efficiency}

Table~\ref{tab:resuts} shows the reported training time of different semi- and unsupervised methods to fully converge. slimIPL has a clear advantage in training time and resource consumption. Among other things, we attribute this to the dynamic pseudo-label cache and the stability of the algorithm.

\subsection{Conversational Speech}
\label{sec:swb}

To analyse applicability of slimIPL and found hyperparameters to other data we perform experiments on the conversational telephone speech, SwitchBoard \& Fisher. 
To create a training set, we combine Switchboard (SB)~\cite{godfrey1993switchboard} (300h) as labeled data and Fisher (FS)~\cite{cieri2004fisher,cieri2004fisher2} (2k hours) as unlabeled data. We use RT-03S~\cite{rt03s} as the validation set; test sets are the Hub5 Eval2000~\cite{hub5} data with two subsets, SwitchBoard (SB) and CallHome (CH). For the data processing and evaluation, we follow the recipe provided by Kaldi~\cite{daniel2011kaldi}. Original sample rate of 8kHz is preserved. We use the same model architecture and optimization strategy as for LibriSpeech experiments from Section~\ref{sec:acoustic}. 
Dropout for supervised baseline is set to 0.3 and decreased to 0.1 during training on unlabeled data.
For slimIPL hyperparameters we use similar values found in LibriSpeech experiments: $C=1000$, $M=20000$, $p=0.1$, $\lambda=1$. For SpecAugment we use the Switchboard setting from~\cite{park2019specaug}: two frequency masks with frequency mask parameter $F=15$, two time masks with time mask parameter $T=70$ and maximum time-mask ratio $p=0.2$; time warping is not used. 
Average batch size is 38 per GPU. 

Fisher data contains samples with silence only. 
From experiments we noticed that these samples used in training can cause instability and slow divergence at the end of training.
Closer to end of training, their negative contribution rises as the model tends to generate empty pseudo-labels more and more frequently. 
After filtering original Fisher data to exclude samples with empty transcriptions, training is stable and no side-effects are observed. 

Table~\ref{tab:resuts-swb} compares results for supervised baseline, slimIPL and RASR~\cite{likhomanenko2020rethinking} (supervised baseline trained on both SB and FS), which share the same model architecture except for dropout rates.
slimIPL outperforms fully supervised baseline (RASR) on RT-03S and in the same ballpark for Hub5 Eval2000 SwitchBoard set.
Thus, we conclude that slimIPL and its hyperparameters are readily applicable to conversational speech.

\begin{table*}[t!]
\caption{
Comparison between supervised and slimIPL models on SwitchBoard.
\label{tab:resuts-swb}}
\begin{footnotesize}
\begin{center}
\setlength\tabcolsep{5pt} 
\begin{tabular}{@{}ccccccccccc@{}}
\toprule
\multirow{2}{*}{Method} & \multirow{2}{*}{Sup. Data} & \multirow{2}{*}{Unsup. Data} & \multirow{2}{*}{Stride} & \multirow{2}{*}{Tokens} & \multirow{2}{*}{Criterion} & \multirow{2}{*}{LM} & Dev WER & \multicolumn{2}{c}{Test WER} \\
\cmidrule(lr){9-10}
 & & & & & & & RT03S        & SB        & FS  \\
\midrule
\multirow{2}{*}{Our baseline} & \multirow{2}{*}{SB} &  \multirow{2}{*}{-} & \multirow{2}{*}{30ms} & \multirow{2}{*}{letters} & \multirow{2}{*}{CTC} & - & 17.5 & 9.6 & 19.3 \\
  & & & & & & word 4-gram & 14.5 & 8.0 & 16.7 \\
\midrule
 \multirow{2}{*}{RASR~\cite{likhomanenko2020rethinking}} & \multirow{2}{*}{SB+FS} &  \multirow{2}{*}{-} & \multirow{2}{*}{30 ms} & \multirow{2}{*}{letters} & \multirow{2}{*}{CTC} & - & 12.0 & 6.9 & 11.4 \\
    & & & & & & word 4-gram & 10.4 & 6.5 & 10.3 \\
  \midrule
\multirow{2}{*}{slimIPL} &  \multirow{2}{*}{SB} & \multirow{2}{*}{FS} & \multirow{2}{*}{30ms} & \multirow{2}{*}{letters} & \multirow{2}{*}{CTC} & - & 11.8 & 7.9 & 14.2 \\
   & & & & & & word 4-gram & 9.6 & 6.8 & 12.1 \\
 \bottomrule
\end{tabular}
\end{center}
\end{footnotesize}
\end{table*}

\section{Conclusion}
We revisit a key component of recent pseudo-labeling success in ASR, beam-search decoding with an LM, and propose \papername{}, in which a single model iteratively re-generates hard pseudo-labels with a dynamic cache to stabilize optimization. slimIPL is robust to hyperparameter changes and substantially simplifies training compared to other semi/unsupervised approaches, while delivering competitive performance for low-resource settings on \librispeech{} test sets. For inference, \papername{} is less prone to LM over-fitting than methods which use an LM for PL generation.

\section{Acknowledgement}
We thank Alex Rogozhnikov for insightful discussions about the algorithm and experiments, and Gil Keren for results discussions.

\bibliographystyle{IEEEtran}

\bibliography{mybib}

\end{document}